\documentclass[onecolumn,journal,11pt]{IEEEtran}
\usepackage{algorithm}
\usepackage{algorithmic}
\usepackage{amssymb,amsmath}
\usepackage{graphics}
\usepackage{graphicx}
\usepackage{setspace}
\usepackage{color}

\newcommand{\N}{\cal{N}}
\newcommand{\B}{\cal{B}}
\newcommand{\J}{\cal{J}}
\newcommand{\C}{\cal{C}}
\newcommand{\V}{\cal{V}}

\newcommand{\okreview}[1]{\noindent{#1}}

\ifCLASSINFOpdf
\else
\fi
\begin{document}
%
\title{Basis Identification for Automatic Creation of Pronunciation
Lexicon for Proper Names}
%
%
%

\author{Sunil Kumar Kopparapu,~\IEEEmembership{Member,~IEEE,}
        Laxmi Narayana M
\thanks{Sunil Kumar Kopparapu is with TCS Innovation Lab - Mumbai, Tata
Consultancy Services, Yantra Park, Thane (West), Maharastra 400601. Email:
SunilKumar.Kopparapu@TCS.Com}
}

\maketitle

\begin{abstract}
Development of a proper names pronunciation lexicon is usually a manual
effort which can not be avoided. Grapheme to phoneme (G2P) conversion
modules, in literature, are usually rule based and work best for
non-proper names in a particular language. Proper names are foreign to a
G2P module. We follow an optimization approach to enable automatic
construction of proper names pronunciation lexicon. The idea is to
construct a small orthogonal set of words (basis) which can span the set
of names in a given database. We propose two algorithms for the
construction of this basis. The transcription lexicon of all the proper
names in a database can be produced by the manual transcription of only
the small set of basis words. We first construct a cost function and
show that the minimization of the cost function results in a basis. We
derive conditions for convergence of this cost function and validate
them experimentally on a very large proper name database.
Experiments show the transcription can be achieved by transcribing a set
of small number of basis words. The algorithms proposed are generic and
independent of language; however performance is better if the proper
names have same origin, namely, same language or geographical region.
\end{abstract}


%
\IEEEpeerreviewmaketitle

\section{Introduction}
%
%
%
%
%
%

bel{sec:introduction}
Text to Speech (TTS) synthesis is an automated encoding process which converts
text (a sequence of symbols
conveying linguistic information), into speech (an acoustic waveform).
The two major components of a TTS synthesizer are (a) natural language
processing (NLP) module,
which produces a phonetic transcription of the given text and (b) digital
signal processing module,
which transforms sequence of phones into speech \cite{dutoit}.
Text normalization is the process of converting non-standard words like
abbreviations, acronyms,
dates, special symbols (for e.g. Dr, Mr, \$700) into their corresponding
graphemic representation \cite{G2P}.
Grapheme to phoneme (G2P) conversion is then performed on the normalized text.
In general, an NLP module should be able to normalize the input text
and map the grapheme representation of the text to a corresponding phonetic
representation.

{A} pronunciation dictionary provides a means to map a word into its
elementary phonetic components
which is a key for modeling TTS synthesis systems.
The reason for this is that in general, a one to one correspondence between the
orthographic representation
of a word and its pronunciation is absent.
However, the need for a pronunciation dictionary reduces by developing a set of
predefined rules (called G2P rules) developed
based on linguistic knowledge, that map a sequence of characters (graphemes)
into a sequence of phones.
The G2P rule base is a set of rules that modify the 'default mapping' of the
characters
based on the 'context' in which a particular phoneme occurs. Specific contexts
are matched using rules.
The system triggers the rule that best fits the current context \cite{G2P}.
G2P converters usually produce a significant number of mistakes when converting
proper names which are often of a foreign origin \cite{P2P}.
The rule set developed is language dependent and hence an existing rule base
for
one language cannot automatically
be used to generate the phonetic transcription of a word from another language.
Proper names being foreign to the G2P rule base of any language, demand manual
effort which
is inevitable to obtain the phonetic transcription. Recently, Bonafonte et al
\cite{pr_Bliz_2007}
reported an average phonetic accuracy of 53\% for proper names
when a rule based methodology is used to construct a phonetic dictionary of
proper names.
Van den Heuvel et al \cite{better} tried to automate the process of
transcribing proper names
by using a cascade of a general purpose G2P converter and a special purpose P2P
(phoneme to phoneme) converter; the P2P converter learns from human expert
knowledge.
Though they report enhanced performance with the cascade system compared to
direct rule based method, the performance of cascade system results in
more than 30\% of the name transcriptions being erroneous. In a manual effort,
Font Llitjos and Black \cite{Evalonline} adopted a web-based interface to
improve pronunciation models as well as correct the pronunciations
in the CMU dictionary by evaluating and collecting proper name pronunciations
online.
Font Llitjos and Black \cite{lang_orgn} \cite{Llitjos01knowledgeof}
hypothesized that higher pronunciation accuracy
can be achieved by adding the knowledge that people adapt their
pronunciation according to where they think a proper name comes from,
to a statistical model of pronunciation.
The ONOMASTICA project \cite{Project95transcribingnames}
\cite{Consortium95theonomastica}, a European wide research initiative, aims at
the construction of
multi-language pronunciation lexicon for proper names by upgrading the existing
rule engines to cope with the problems posed by proper names.
A significant part of the work was also devoted to the development of
self-learning
G2P conversion methods and the comparison of their performance with the
one of rule-based methods.

A general purpose G2P rule base cannot cater to proper names because
such rule bases are developed for a particular language cannot be generalized
to all kinds of words,
especially for the proper names.
This means, there is a need to develop a pronunciation
dictionary for proper names. But the development of such a
lexicon\footnote{We will use lexicon and dictionary interchangeably in this
paper}
is not possible by having a mere rule set; it demands manual effort to generate
phonetic
transcriptions of a large set of names.
A possible solution is to create a small set of
words\footnote{Words could be names themselves or part of names}
which when phonetically transcribed, manually, can span and hence transcribe
all the proper names in a given database. Obviously, the choice of
the words that have to be transcribed should be such that they occur frequently
in the database of names.

Several problems which were solved by constructing a cost function and finding
the extremes
(maxima, minima) are mentioned in literature (for e.g., \cite{pr_Toda1},
\cite{pr_Toda2}, \cite{jr_Toda3})
This paper describes a method to enable construction of this set of
words derived from the actual proper names database. We call it basis in a
loose sense;
taking cue from vector algebra.
We construct a cost function which when minimized results in the identification
of a basis.
This can then be used in phonetic transcription of the full database of proper
names.
The rest of the paper is organized as follows. Section \ref{sec:problem}
formulates the identification of basis as an
optimization problem. We also discuss the trivial cases of creation of a basis
for proper names.
Section 3 describes the proposed algorithms for basis creation of proper names.
Section 4 presents experimental results and we conclude in Section 5 and also
give future directions.

\section{Problem Formulation}
\label{sec:problem}
We address the following problem.

\begin{quote}
\emph{Given a proper name database of $|{\N}|$ names} (\emph{e.g.,}
\{\emph{\texttt{rama, krishna, narayana ...}}\}),
\emph{can we construct a smaller set of words} (\emph{basis, e.g.,}
\{\texttt{\emph{ra, na, krish, ya, ...}}\})
\emph{automatically, such that all the $|{\N}|$ names can be formed by the
words in the basis, namely\\
\\
\texttt{rama = ra $\oplus$ ma}}\footnote{$\oplus$ represents a join}\\
\emph{\texttt{krishna = krish $\oplus$ na\\
narayana = na $\oplus$ ra $\oplus$ ya $\oplus$ na\\
...}}
\end{quote}
Given a database of proper names\footnote{The proper names are written in Roman
script}, two trivial cases of building a
pronunciation lexicon are possible.
(a) At one extreme one could build a pronunciation lexicon by manually
transcribing all the names in the proper names database
and (b) On the other extreme one could have a pronunciation dictionary of the
26 letters of the
English alphabet and use that to construct the pronunciation lexicon of all
the names in the dictionary by concatenating the letters that make the name.
Obviously, the first trivial case is manually intensive while the second
trivial case is
manually easy but introduces as many joins as the number of letters that make
the name;
as a result the pronunciation produced is feeble. The question that one is
posing here is
\emph{"Is there an optimal set of words that one can identify and manually
transcribe
so that it can be used to produce a good pronunciation dictionary?"}.
In other words, is there an optimal set of words such that the need for manual
transcription is \emph{small} and at the same time the pronunciation of the
names
in the database is \emph{good?} In this paper we construct a cost function
which 
helps us achieve a set of words (we call it the basis because it has properties
of a basis)
which can be used to construct the pronunciation dictionary of the full set of
proper names.

We make use of a restricted definition of basis (see Appendix A) to assist our
problem formulation.
In our case, the \emph{vector space} is the complete set of names in the
database and the \emph{basis}
is a set of words such that, one can construct a name in the database by
joining one or more words from the basis.
Further, no word in the basis can be formed by joining one or more words in the
basis (Property 1, Appendix A).
This is analogous to the scenario of concatenative speech synthesis where one
looks for the longest possible speech unit to synthesize speech with minimal
discontinuities.

The optimization required is that the number of entries in the basis should be
as small as possible to minimize the manual effort to transcribe them and at
the 
same time the number of basis words (joins) used to construct a name in the
database should be small.
These two requirements are contradicting and hence the need for optimization.

Let ${\N} = \{{\N}_{1}, {\N}_{2}, ... , {\N}_{|{\N}|}\}$ represent all the
names 
in the proper names database and let ${\B} = \{b_{1}, b_{2}, ..., b_{|{\B}|}\}$
be the basis
satisfying the linear independence property of Appendix A, namely
for every $b_{k} \in {{\B}}; b_{k}$ can not be expressed as
$b_{i} \oplus b_{j} \oplus ... \oplus b_{l}$,
using any $b_{i}, b_{j}, ..., b_{l} \in {\B}$ and $b_{i}$ or $b_{j}$ or ... or
$b_{l} \neq b_{k}$.
This implies $b_{1} \perp b_{2} \perp ... \perp b_{|{\B}|}$. Additionally, for
any name ${\N}_{p} \in {\N}$, one can write

\begin{equation}
{\N}_{p} = \oplus^{n}b_{i} \qquad b_{i} \in {{\B}}
\end{equation}
This is equivalent to saying that ${\N}_{p}$ can be represented by a join of
some $n$
elements in the basis set ${\B}$ which results in ${\J}_{p} = (n-1)$ joins.
Thus the total number of joins, $|{\J}|$ required to construct the entire
database of $|{\N}|$ names is

\begin{equation}
|{\J}| = \sum_{p=1}^{|{\N}|}{\J}_{p}
\end{equation}

\subsection{Trivial cases}
As mentioned earlier, two trivial cases of construction of pronunciation
dictionary are possible.\\
\emph{Case (i)}: If the number of joins to construct the names is to be small
then
all the names in the database should be present in the basis set and this would
result in the largest basis, say ${\B}_{max}$ which would have all the $|{\N}|$
names in the database.
Further there is a possibility that ${\B}_{max}$ is not a basis in the sense
defined in Appendix A.
This is shown in Figure \ref{fig:tcase1}.
\begin{figure}

\vbox{
\begin{center}
${\N}_{1} = b_{1}$\\
${\N}_{2} = b_{2}$\\
${\N}_{3} = b_{3}$\\
.\\.\\.\\
${\N}_{|{\N}|}=b_{|{\N}|}$\\
.\\
${\B}={\N}; |{\B}|=|{\N}|; |{\J}|=0$
\end{center}
}
\caption{Trivial \emph{Case (i)} where $|{\B}|=|{\N}|; |{\J}|=0$ }
\label{fig:tcase1}
\end{figure}
\\\emph{Case (ii)}: The smallest possible basis ${\B}_{min}$ would be the set
of 26
letters in the English alphabet and this basis would definitely span the entire
database of names,
but the number of joins, $|{\J}|$ required to form the names in the database
would be very large.
This is shown in Figure \ref{fig:tcase2}.
\begin{figure}
\vbox{
\begin{center}
${\N}_{1} = \alpha_{1}\oplus\alpha_{25}\oplus\alpha_{5}$\\
${\N}_{2} = \alpha_{1}\oplus\alpha_{12}\oplus\alpha_{3}$\\
${\N}_{3} = \alpha_{11}\oplus\alpha_{15}\oplus\alpha_{17}\oplus\alpha_{29}$\\
.\\.\\.\\
${\N}_{|{\N}|} = \alpha_{14}\oplus\alpha_{20}$\\
.\\
${\B}=\{\alpha_{i}\}_{i=1}^{26}; |{\B}|=26; |{\J}|\approx\infty$
\end{center}
}
\caption{Trivial \emph{Case (ii)} where $|{\B}|=26$ and $|{\J}|\approx\infty$}
\label{fig:tcase2}
\end{figure}

\begin{figure}
\centering
\includegraphics[width=0.30\textwidth]{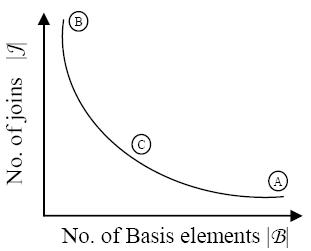}
\caption{Plot between number of elements in the basis, $|{\B}|$ and the total
number of joins $|{\J}|$}
\label{fig:BJ}
\end{figure}

\begin{figure}
\centering
\includegraphics[width=0.30\textwidth]{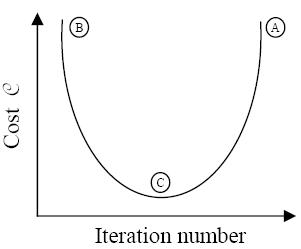}
\caption{Cost (${\C}$) of constructing the pronunciation dictionary}
\label{fig:globalcost}
\end{figure}

A typical plot of the number of elements in the basis $|{\B}|$ versus the total
number of joins required to construct all the names in the database $|{\J}|$,
is shown in Figure \ref{fig:BJ}
. The scenario depicted in \emph{Case (i)} corresponds to the point
A in Figure \ref{fig:BJ}
and the \emph{Case (ii)} corresponds to the point B in Figure \ref{fig:BJ}.
We believe that the cost of construction of basis would be maximum at these two
extreme trivial cases.
Probably there is a case between these two trivial solutions; like the knee
point C
at which the cost of construction of the basis would be minimum as
shown in Figure \ref{fig:BJ}
and \ref{fig:globalcost}
which can be achieved. We investigate if we can identify C in Figures
\ref{fig:BJ}
and \ref{fig:globalcost}.
One has the choice of identifying the basis by starting from an initial basis.
There are 4 different ways of initializing the basis.
(a) start at point A, (b) start at point B, (c) choose some ${\B}_{init}$ and
(d) start with ${\B}$ = null.
We experiment with cases (c) and (d). Note that in both of these cases, we are
traversing through only a portion
of the curves shown in Figures \ref{fig:BJ}
and \ref{fig:globalcost}
meaning starting at some point on the curve
and reaching the knee point C.

Let ${\B}_{init}$ be an initial basis. Then a name
$\{{\N}_{p}\}_{p=1}^{|{\N}|}$ in the database can, in \emph{Case (a)},
be completely represented by using some of the elements in the basis,
namely, ${\N}_{p} = b_{i} \oplus b_{l} \oplus ... \oplus b_{m}$,
where $b_{i}, b_{l}, ..., b_{m} \in {\B}_{init}$ and in \emph{Case (b)} be
partially
represented, namely, ${\N}_{p} = b_{i}\oplus n_{b} \oplus b_{m}$,
where $b_{i}, b_{m} \in {\B}_{init}$, and $n_{b} \notin {\B}_{init}$.
In \emph{Case (b)}, for ${\N}_{p}$ to be representable using the basis,
we need to necessarily add $n_{b}$ to ${\B}_{init}$ and further make sure
that $n_{b}$ is orthogonal to all the elements in ${\B}_{init}$,
namely $\{b_{i}\}_{i=1}^{|{\B}_{init}|}$. The addition of $n_{b}$ introduces an
extra element into ${\B}_{init}$, hence increasing size of the basis
$|{\B}_{init}|+1$. In reality we need to keep the size of basis as small as
possible.
The identification of an optimal basis set reduces to an optimization problem.
Specifically, to optimize a function of $|{\B}|$ and $|{\J}|$. Namely,\\
\begin{equation}
\label{eq:cost}
{\C} = {\LARGE f}(|{\B}|,|{\J}|) = |{\B}|\left\{1+\frac{|{\J}|}{|{\N}|}
\right\}
\end{equation}
where ${\C}$ is the cost of construction of the optimal basis ${\B}$.
Figure 4 shows the variation of the cost function ${\C}$ for different
combinations
of $|{\B}|$\footnote{Here, $|{\B}|$ is not a basis in a strict sense as defined
in Appendix A}
and $|{\J}|$. 
constructs the basis. This will be clearer in Section 4. 
Now the optimization problem can be stated as\\
\begin{equation}
\min_{{\B}}\{{\C}\}
\end{equation}
meaning, choose a basis ${\B}$ such that ${\C}$ is minimized.
The object is to find the knee point C
at which the cost (${\C}$) of construction of the
optimal basis (${\B}$) corresponding to which the number of joins (${\J}$)
are "reasonable", would be minimum. The cost ${\C}$, would be maximum at the
two extreme points A and B in Figures 3 and 4.
At point A, $|{\B}|=26$ and $|{\J}|=\sum_{i=0}^{|{\N}|} \left (l_{i}-1 \right
)$ 
where $l_{i}$ is the length of the name ${\N}_{i}$. In this case, as $|{\N}|
\to \infty, |{\J}| \to \infty$ and hence ${\C} \to \infty$.
At point B, $|{\B}|=|{\N}|$ and $|{\J}|=0$. In this case, as $|{\N}| \to
\infty$, ${\C} \to \infty$.

This formulation seeks the construction of an optimum basis that can span the
entire
database which can be achieved with optimal values for the two parameters
$|{\B}|$ and $|{\J}|$ together.
The expectation is that the optimum basis is created at some knee point C on
the 
curve shown in Figures 3 and 4, where the number of basis elements and the
number of joins are optimal.
\section{Algorithms for the construction of Basis}
We propose two algorithms for the construction of the basis -
one with a choice of initial basis (see Algorithm 1) and the other with out an
initial basis (see Algorithm 2).

\emph{Definitions}
\begin{itemize}
\item Rank deficient basis: The basis set is called \emph{rank deficient} if it
is non-orthogonal meaning,
some of the members in the set can be constructed using other entries in the
set.
A rank deficient basis does not satisfy the linear independence property
(Appendix A).
\item Span deficient basis: The basis set is called \emph{span deficient} if it
does not span
the entire proper names database meaning,
all the names in the database can not be constructed using this set.
A span deficient basis does not satisfy the spanning property (Appendix A).
\end{itemize}
\subsection{Algorithm 1}

\begin{algorithm}[t]
\begin{algorithmic}
\STATE{${\N} = \{{\N}_{1}, {\N}_{2}, ..., {\N}_{p}, ..., {\N}_{|{\N}|}\}$}
\STATE{1. \texttt{initialize} ${\B}_{init}$ = $\{b_{1}$, $b_{2}$, $\cdots$,
$b_{i}$, $\cdots$, $b_{|{\B}_{init}|}\}$}
\STATE{2.  \texttt{isOrtho}(${\B}_{init}$)}
\STATE{3.  ${\B}_{init}$ = \texttt{makeOrtho}(${\B}_{init}$)}
\STATE{4. \textbf{\texttt{do}}}
\STATE{5.   \qquad \texttt{initialize} ${\B}_{m} = {\B}_{init}$ }
\STATE{6.   \qquad \textbf{\texttt{for}} \texttt{each name} ${\N}_{p} \in {\N}$
}
\STATE{7.     \qquad \{\texttt{form} ${\B}_{p} = \{b_{p1}, b_{p2}, ...,b_{pk},
..., b_{p|{\B}_{p}|}\}$}
\STATE{       \qquad \qquad    \texttt{such that}  ${\B}_{p} \subset
{\B}_{init}$ and}
\STATE{          \qquad  \qquad        $b_{pk}$ \texttt{is a substring of}
${\N}_{p}$ }
\STATE{8.     \qquad \texttt{Identify all possible\_sequences of} }
\STATE{         \qquad \qquad ${\N}_{p}$ \texttt{using} ${\B}_{p}$
\{${\N}_{p1}, {\N}_{p2}, ...,{\N}_{pl}, ..., {\N}_{pr}$\}}
\STATE{9.     \qquad \texttt{collect all the new words}}
\STATE{10.     \qquad \textbf{\texttt{for}} \texttt{each} ${\N}_{pl}$,}
\STATE{        \qquad\qquad \texttt{obtain\_cost} $C_{pl}$ of ${\N}_{pl}$}
(Equation 16)
\STATE{      \qquad \qquad \textbf{\texttt{end for}}}
\STATE{11.   \qquad \texttt{choose} ${\N}_{pl}$ \texttt{with minimum} $C_{pl}$
}
\STATE{    \qquad\qquad  \texttt{add new words in} ${\N}_{pl}$ to ${\B}_{m}$}
\STATE{ \qquad \qquad \textbf{\texttt{end for}}}
\STATE{12.  \qquad \texttt{isOrtho}(${\B}_{m}$)}
\STATE{13.  \qquad${\B}_{init}$ = \texttt{makeOrtho}(${\B}_{m}$)}
\STATE{14.  \qquad\texttt{goto step} 5}
\STATE{15. \textbf{\texttt{until}}$(|{\B}_{m}|-|{\B}_{init}|)<\epsilon)$}
\STATE{16. ${\B}_{opt}={\B}_{init}$}
\end{algorithmic}
\caption{Pseudo-code}
\label{Algo:Pseudo_code}
\end{algorithm}

\noindent \textbf{Step 1, \texttt{initialize} ${\B}_{init}$}: The basis is
initialized by sorting the $|{\N}|$ names
in the database in the descending order of the number of occurrences in the
database and then picking up
all the names whose frequency of occurrence is greater than or equal
to k\% 
of the
maximum frequency\footnote{'Maximum frequency' refers to the frequency of the
name which occurs
the most number of times in the database.}.

\noindent \textbf{Step 2, \texttt{isOrtho(${\B}_{init}$)}}: ${\B}_{init}$ is
checked for its orthogonality,
namely, it is checked if any word in it can be completely
constructed with a combination of other words in it. This task is accomplished
by \texttt{isOrtho()}.

\noindent \textbf{Step 3, \texttt{makeOrtho(${\B}_{init}$)}}: If ${\B}_{init}$
is found to be rank deficient (non-orthogonal),
there is a need to make it orthogonal. If an element is found to be completely
constructed with
other elements in the set ${\B}_{init}$, that element is deleted from the set.
This task is accomplished by a function named \texttt{makeOrtho()}.
The process of orthogonalization of basis is described briefly in Appendix C.

\noindent \textbf{Step 4}: Start an iteration of constructing the basis.

\noindent \textbf{Step 5, \texttt{initialize} ${\B}_{m}$ = ${\B}_{init}$}:
We initialize a new set ${\B}_{m}$ with ${\B}_{init}$.
${\B}_{m}$ will be used to store the new words (that are not in ${\B}_{init}$)
required to construct all the names in ${\N}$ if ${\B}_{init}$ is span
deficient.

\noindent \textbf{Step 6, \texttt{for each name} ${\N}_{p} \in {\N}$}, we do
the following.

\noindent \textbf{Step 7, \texttt{forming} ${\B}_{p}$}: Let ${\B}_{p} =
\{b_{p1}, b_{p2}, ... , b_{|{\B}_{p}|}\}$
such that ${\B}_{p} \in {\B}$ and can completely or partially construct the
name ${\N}_{p}$ in the database,
namely, all the basis words in ${\B}_{p}$ are substrings\footnote{We consider a
word
as a substring of a name if it is a part of the name or sometimes the name
itself}
of the name ${\N}_{p}$. Note that ${\B}_{p}$ can be a null set meaning there
are no
elements in the basis which is a substring of ${\N}_{p}$.
In such a case ${\N}_{p}$ should be added to the basis.

\noindent \textbf{Step 8, construction of \texttt{possible sequences}
(${\N}_{p}$) \texttt{with} ${\B}_{p}$}:
Consider ${\B}_{p}$ is not empty; then the words in ${\B}_{p}$ may partially or
completely construct
${\N}_{p}$ (see Equation (1)). Let the name ${\N}_{p}$ be constructed with
${\B}_{p}$ in $r$
different ways as shown in Figure 5.

\begin{figure}
\fbox{
\vbox{
${\B} = {b_{1}, b_{2}, ... , b_{|{\B}|}}$\\
${\B}_{p} = {b_{p1}, b_{p2}, ... , b_{|{\B}_{p}|}}$\\
\\
${\N}_{p1} = b_{p3} \oplus b_{p5} \oplus b_{p2}$\\
${\N}_{p2} = n_{p21} \oplus b_{|{\B}_{p}|}$\\
.\\.\\.\\
${\N}_{pl} = b_{p4} \oplus n_{p31} \oplus b_{p7} \oplus n_{p32}$
\\.\\.\\.\\
${\N}_{pr} = b_{p5} \oplus b_{p6} \oplus b_{p4}$\\
where\\
${\N}_{p1} = {\N}_{p2} = ... = {\N}_{pl} = ... = {\N}_{pr} = {\N}_{p}$
}
}
\caption{Different ways of constructing ${\N}_{p}$ with ${\B}_{p}$}
\end{figure}
As seen in Figure 5 there are $r$ possible sequences constructed for the name
${\N}_{p}$.
The choice of the $l^{th}$ sequence is represented as ${\N}_{pl}$. For example,
the sequence
${\N}_{p2}$ requires a new word $n_{p21}$ which is not in ${\B}_{p}$,
for successfully constructing ${\N}_{p}$, while ${\N}_{pl}$ requires two new
words
$n_{p31}$ and $n_{p32}$ to construct ${\N}_{p}$, while ${\N}_{p1}$ and
${\N}_{pr}$
completely construct the name without the aid of any new word being added to
the existing basis.
If more than one such representation of ${\N}_{p}$ is possible using different
combination of words
in ${\B}_{p}$, then a decision has to be taken as to which representation is to
be retained.
In such case, the selection depends on the cost of constructing the sequence.

Some sequences might partially construct the name.
Or sometimes, none of the sequences might completely construct the name.
In the later case, there is a need to include some new words into the basis,
to enable the basis to construct the name (and the entry should also be
orthogonal to the existing basis as mentioned earlier).
So, a decision has to be taken about which new word(s)
should be added to the basis and what is the cost of such addition.

\noindent \textbf{Step 9, \texttt{collect all the new words}}: The new words
required by all
the sequences formed for a name are collected and their frequency of occurrence
is calculated.
Note that even if more than one sequence formed for a name require a new word,
the new word's frequency is counted only once.
Thus the maximum value of frequency for a new word would be $|{\N}|$ meaning
that this
new word is required by all the names in the proper name database.

For every name in the database, we have several possible sequences and a list
of words which are not in the basis.
We need to choose one of the $r$ sequence choices to represent the name
${\N}_{p}$.
The choice is one that results in (a) minimal number of joins and (b) adds
minimal number of
entries to the existing basis set (${\B}_{init}$).
Observe that there is a need for optimality in choosing one of the $r$
sequences.
We construct a cost function to identify the optimal sequence choice.

\noindent \textbf{Step 10, \texttt{obtaining cost for each word sequence}
${\N}_{pl}$}:
Let the $l^{th}$ sequence (${\N}_{pl}$) out of the $r$ sequences that represent
${\N}_{p}$ has $\eta_{pl}$ number of words, \{$s_{1},
s_{2},...,s_{k},...,s_{n}$\}
out of which $\eta_{new}$ are new and $\eta_{ex}$ belong to the existing basis,
meaning
$\eta_{pl} = \eta_{new} + \eta_{ex}$. If $\eta_{joins}$ is the number of joins
in the $l^{th}$ sequence, then
\begin{equation}
\eta_{joins} = (\eta_{pl}-1)
\end{equation}
Let $L$ be the length (number of letters) of the name ${\N}_{p}$.
The cost function $C_{pl}$ is formulated as
\begin{equation}
C_{pl}=f_{pl}(\mu,\nu,\eta_{pl},\eta_{new},\eta_{joins},P_{av},F_{av},SA_{av})
\end{equation}
a function of the parameters of the word sequence ${\N}_{pl}$
namely, $\mu$, $\nu$, $\eta_{joins}$, $\eta_{new}$, $\eta_{pl}$, $P_{av}$,
$F_{av}$ and $SA_{av}$.
And we choose the sequence ${\N}_{pl}$ such that
\begin{equation}
{\N}_{pl} = \underset{(l)}{\operatorname{argmin}}\big\{C_{pl}\big\}\qquad{1
\leq l \leq r}
\end{equation}
The cost function $C_{pl}$ is best described by looking at each element
involved in the construction of $C_{pl}$.
We identify the relevance of the features and the redundancy in them in the
following discussion.
$\mu$ is the average length of the words in a sequence which is given by
\begin{equation}
\mu = \frac{1}{\eta_{pl}}\sum_{k=1}^{n_{pl}}l_{k}
\end{equation}
where $l_{k}$ is the length of the $k^{th}$ word $s_{k}$ in ${\N}_{pl}$.
Maximization of $\mu$ reduces the number of joins in the sequence.
Observe that the component $\sum_{k=1}^{\eta_{pl}}l_{k} = L$.
Hence, Equation (8) reduces to $\frac{\mu}{L} = \frac{1}{\eta_{pl}}$.
So, maximization of $\mu$ means minimization of $\eta_{pl}$ (number of words in
the sequence)
which in turn reduces the number of joins, $\eta_{joins}$ (see Equation (5)) in
the sequence ${\N}_{pl}$.
Hence, considering one of these three parameters is sufficient in formulating
the cost function.
$\nu$ is the variance of the lengths of words in the sequence ${\N}_{pl}$ and
is given by
\begin{equation}
\nu = \frac{1}{\eta_{pl}}\sum_{k=1}^{\eta_{pl}}(l_{k}-\mu)^{2}
\end{equation}
$P_{av}$ is the average percentage of acceptance of the words in the sequence
${\N}_{pl}$ and is given by
\begin{equation}
P_{av} = \frac{1}{\eta_{pl}}\sum_{k=1}^{\eta_{pl}}pd_{k}
\end{equation}
where $pd_{k}$ is the 'percentage demand' of the word from all the $r$
sequences formed for the present name
namely, $pd_{k}$ is the percentage of $r$ sequences formed for ${\N}_{p}$ which
require $s_{k}$.
\begin{equation}
pd_{k} = \frac{\textrm{\emph{Number of sequences demanding }}s_{k}}{r}
\end{equation}
$F_{av}$ which is defined only for the new words in the sequence, is the
average frequency
of occurrence of the new words in ${\N}_{pl}$ and is given by
\begin{equation}
F_{av} = \frac{1}{\eta_{new}}\sum_{k=1}^{\eta_{new}}f_{k}
\end{equation}
where $f_{k}$ is the frequency of occurrence of the new word $s_{k}$ as a basis
element,
namely, $f_{k}$ is the percentage of names in the database that are in
requirement of $s_{k}$
for their construction (even if one of the $r$ sequences formed for ${\N}_{p}$
requires the word).
So, $f_{k}$ is given by
\begin{equation}
f_{k} = \frac{\textrm{\emph{Number of names requiring }}s_{k}}{|{\N}|}
\end{equation}
$f_{k}$ of every new word is obtained in Step 9.
$SA_{av}$ is a binary valued attribute named by 'Syntax rule acceptance' and is
defined for the new words in the sequence and checks if the word $s_{k}$ to be
introduced into
the basis follows the syntactic rules given in Appendix B.
$sa_{k}$ is set to $1$ if $s_{k}$ follows the syntax rules and is set to $0$ if
it violates the syntax rules.
$\eta_{ak}$ is the number of words following the syntactic rules out of the
$\eta_{new}$
number of new words in the sequence ${\N}_{pl}$ (while the remaining are
violating) and is given by
\begin{equation}
\eta_{ak} = \sum_{j=1}^{\eta_{new}}{sa_{k}}
\end{equation}
$SA_{av}$ is the percentage of new words following the syntactic rules given in
Appendix B and is given by
\begin{equation}
SA_{av} = \frac{\eta_{ak}}{\eta_{new}}
\end{equation}
Ideally, for any word in the sequence, the features $l_{k}$ and $pd_{k}$ should
be
maximum and for a new word to be included into the basis, its $f_{k}$ should be
maximum and $sa_{k}$ should be 1.
We saw earlier that considering one of the three features $\mu$, $n_{pl}$ and
$n_{joins}$ is sufficient.
This implies that the 4 parameters $\mu$, $P_{av}$, $F_{av}$ and $SA_{av}$
defined for a word sequence should be maximized.
In addition, the overall variance ($\nu$) of the lengths of the words and the
number of new words ($\eta_{new}$) in the sequence
should be minimized for the sequence to be optimal.
In other words, the proportionality of the cost of constructing a name
(cost of selecting one of the $r$ sequences formed for a name) with the
features of the word sequence is given as follows
\begin{align*}
C_{pl} &\propto \eta_{pl}\\
&\propto \eta_{joins} \\
&\propto \eta_{new} \\
&\propto \nu\\%
&\propto \frac{1}\mu\\
&\propto \frac{1}{P_{av}}\\%
&\propto \frac{1}{F_{av}}\\%
&\propto \frac{1}{SA_{av}}
\end{align*}
Considering the redundancy in features and their relation with the cost of
construction of a name, we write the function $f_{pl}$ as shown in Equation
(15).
\begin{eqnarray}
\nonumber
&f_{pl}(\mu,\nu,\eta_{new},P_{av},F_{av},SA_{av}) =
\frac{\lambda_{\mu}}{\mu}+\lambda_{\nu}\nu \\
&+\lambda_{p}P_{av}+\lambda_{\eta}\eta_{new}(\frac{1}{F_{av}}+\frac{1}{SA_{av}})
\end{eqnarray}

where $\lambda_{\mu}$, $\lambda_{\nu}$, $\lambda_{p}$ and $\lambda_{\eta}$ are
the weights assigned to $\mu$, $\nu$, $P_{av}$ and the new words of the
sequence respectively such that
\begin{equation}
\lambda_{\mu}+\lambda_{\nu}+\lambda_{p}+\lambda_{\eta} = 1
\end{equation}
We define the weight set as $\Lambda_{1} = \{\lambda_{\mu}, \lambda_{\nu},
\lambda_{p}, \lambda_{\eta}\}$.
\okreview{Note that different choices of $\Lambda_{1}$ result in different
basis sets.
We choose that set which gives the minimal cost.}

\noindent \textbf{Step 11, \texttt{choose} ${\N}_{pl}$ \texttt{with minimum}
$C_{pl}$}:
One sequence among the $r$ sequences (formed for a name ${\N}_{p}$) which gives
the minimum cost is selected (recall Equation (7)) and the new words, if any,
present in the sequence, are stored separately (${\B}_{m}$ in the pseudo code).
This process is repeated for all the proper names in the database.
After all the names in the database are constructed with the existing basis,
we are left with a set of new words to be introduced into the basis.
We also have the frequency of occurrence of each of the new words and which
database name is in requirement of a new word.

In summary, for a given name, the list of candidates from the existing basis
that can construct the present name is collected and the sequences which
partially
or completely construct the name are formed. Based on the cost function
formulated,
one of the sequences that represent the name is selected and new entries,
are made into the existing basis if required.

\noindent \textbf{Step 12, \texttt{isOrtho(${\B}_{m}$)}}: After adding new
elements to the existing basis
(We add all the new words to ${\B}_{m}$ which is initialized to ${\B}_{init}$ -
see Step 5),
(${\B}_{m}$) is checked for its orthogonality (rank deficiency).

\noindent \textbf{Step 13, \texttt{makeOrtho(${\B}_{m}$)}}: If ${\B}_{m}$ is
found to be rank deficient,
it is made orthogonal using the function \texttt{makeOrtho()}.
\begin{quote}
By constructing the names in ${\N}$ with the existing basis (not in a strict
sense),
we check its \emph{spanning property} and if it is found to be span deficient,
by minimizing the cost function,
we add to it, the required words to make it span the entire database.
Then we check for its \emph{orthogonality property} using \texttt{isOrtho()}
and make it orthogonal using \texttt{makeOrtho()}.
\end{quote}
This completes one iteration.

\noindent \textbf{Step 14, \texttt{goto step} 5}:
Once, an orthogonal basis is formed, the database of names are again
constructed with the updated basis.
In this iteration, if some database names are not completely constructed with
the pruned basis,
some new entries are again made in to the basis based on the cost function
formulated.
The new basis is again checked for its orthogonality and pruned if necessary.
The procedure of constructing the names of the database
with the pruned basis is repeated again.
New entries are appended to the basis if required.

\noindent \textbf{Step 15,
\textbf{\texttt{until}}$(|{\B}_{m}|-|{\B}_{init}|)<\epsilon$}:
The process of growing and pruning of the basis (checking for the spanning and
orthogonality properties of the loosely defined basis set)
is stopped when no significant growth and redundancy in the basis are observed
in successive iterations.
Note that $\epsilon$ is a small positive value.

\noindent \textbf{Step 16, ${\B}={\B}_{init}$}: The optimum basis for the
generation of pronunciation dictionary
for the set of proper names, is the set of words obtained in the last iteration
of pruning of the basis.
\subsubsection{\textbf{Conditions for convergence of the cost function}}
We saw in Section 2 that the cost of constructing the basis is maximum at the
points A and B in
Figure 4 and an optimal basis is achieved at the knee point C.
If $|{\J}_{init}|$ is the number of joins corresponding to the initial basis
$|{\B}_{init}|$, then
($|{\B}_{init}|$, $|{\J}_{init}|$) is a point between the points A and C or B
and C on the curve shown in Figure 3.
The optimal basis ${\B}_{opt}$ is achieved at Step 14 in Algorithm 1 where the
cost function converges to the knee point C.
Note that in Algorithm 1, ${\C}$ is non increasing.
Let the cost ${\C}$ at $n^{th}$ iteration be ${\C}_{n}$ and at $(n+1)^{th}$
iteration be
${\C}_{n+1}$, then
\begin{eqnarray*}
{\C}_{n} &=& |{\B}_{n}|\left \{1+\frac{|{\J}_{n}|}{|{\N}|} \right \}  \\
{\C}_{n+1} &=& |{\B}_{n+1}|\left \{1+\frac{|{\J}_{n+1}|}{|{\N}|} \right \}.  
\end{eqnarray*}
Convergence of the cost function is achieved when ${\C}_{n+1} \le {\C}_{n}$ or
${\C}_{n} - {\C}_{n+1} \ge 0$ which reduces to
\begin{eqnarray*}
&&\hspace{-8mm}(|{\B}_{n}|-|{\B}_{n+1}|)+\\
&&\frac{1}{|{\N}|} \left ( |{\B}_{n}||{\J}_{n}| - |{\B}_{n+1}||{\J}_{n+1}|
\right ) \ge 0.
\end{eqnarray*}
The cost function is convergent if and only if the conditions (\ref{cond1}) and
(\ref{cond2}) are satisfied.
\begin{eqnarray}
\label{cond1}
|{\B}_{n+1}| &\le& |{\B}_{n}| \\
|{\B}_{n+1}||{\J}_{n+1}| &\le& |{\B}_{n}||{\J}_{n}|
\label{cond2}
\end{eqnarray}
The convergence is validated through experiments in Section 4.

ubsection{Algorithm 2}
\begin{algorithm}
\begin{algorithmic}
\STATE{${\N} = \{{\N}_{1}, {\N}_{2}, ..., {\N}_{p}, ..., {\N}_{|{\N}|}\}$}
\STATE{${\B} = \{\}$}
\STATE{1.   \textbf{\texttt{for}} \texttt{each name} ${\N}_{p} \in {\N}$ }
\STATE{2.     \qquad \texttt{form all possible\_sequences of} ${\N}_{p}$ }
\STATE{         \qquad \qquad $\{{\N}_{p1}, {\N}_{p2}, ...,{\N}_{pl}, ...,
{\N}_{pr}\}$}
\STATE{3.     \qquad \textbf{\texttt{for}} \texttt{each} ${\N}_{pl}$,}
\STATE{        \qquad\qquad \texttt{obtain\_cost} $C_{pl}$ of ${\N}_{pl}$}
\STATE{      \qquad \qquad \textbf{\texttt{end for}}}
\STATE{4.   \qquad \texttt{choose} ${\N}_{pl}$ \texttt{with minimum} $C_{pl}$ }
\STATE{    \qquad\qquad  \texttt{add new words in} ${\N}_{pl}$ to ${\B}$}
\STATE{5.  \texttt{isOrtho}(${\B}$)}
\STATE{6.  ${\B}_{opt}$ = \texttt{makeOrtho}(${\B}$)}
\STATE{ \qquad \textbf{\texttt{end for}}}
\end{algorithmic}
\caption{Pseudo-code}
\label{Algo2}
\end{algorithm}

\noindent \textbf{Step 1, \texttt{for each name ${\N}_{p} \in {\N}$}} we do the
following.

\noindent \textbf{Step 2, construction of all \texttt{possible sequences}
(${\N}_{p}$)}:

Here, we have no initial basis. We construct the name ${\N}_{p}$ in all
possible ways
in which it can be constructed with its substrings\footnote{substring is a part
of the name}.
An example is possible sequences of the name '\texttt{gopal}' are
\{\texttt{g opal,
go pal,
gop al,
gopa l,
g o pal,
g op al,
g opa l,
go p al,
go pa l,
gop a l,
g o p al,
g o pa l,
g op a l,
go p a l,
g o p a l}\}.
The sequences that have a single letter are excluded from this list.

\noindent \textbf{Step 3, \texttt{obtaining cost for each word sequence}
${\N}_{pl}$}:
In this case, the cost function $C_{pl}$ is based on the parameters $\mu$,
$\nu$, $P_{av}$
and $SA_{av}$\footnote{The definitions of the parameters remain same as
discussed in Algorithm 1.}
and the function $f_{pl}$ is given by
\begin{eqnarray}
\nonumber
&&\hspace{-12mm}f_{pl}(\mu,\nu,P_{av},SA_{av}) \\
&&=
\frac{\lambda_{\mu}}{\mu}+\lambda_{\nu}\nu+\lambda_{p}P_{av}+\frac{\lambda_{s}}{SA_{av}}
\end{eqnarray}
where $\lambda_{\mu}$, $\lambda_{\nu}$, $\lambda_{p}$ and $\lambda_{s}$ are the
weights assigned to $\mu$, $\nu$, $P_{av}$ and the syntax of the words
respectively such that
\begin{equation}
\lambda_{\mu}+\lambda_{\nu}+\lambda_{p}+\lambda_{s} = 1.
\end{equation}
We define the weight set as $\Lambda_{2} = \{\lambda_{\mu}, \lambda_{\nu},
\lambda_{p}, \lambda_{s}\}$.

\noindent \textbf{Step 4, \texttt{choose} ${\N}_{pl}$ \texttt{with minimum}
$C_{pl}$}:
One sequence among the $r$ sequences (formed for a name ${\N}_{p}$) which gives
the minimum cost is selected (recall Equation (7)) and the words in the
sequence 
are added to the basis. This process is repeated for all the proper names in
the database.

\noindent \textbf{Step 5, 6}: The basis thus formed is then checked for its
orthogonality and made
orthogonal using the functions \texttt{isOrtho()} and \texttt{makeOrtho()}.
\section{Experimental Results and Discussion}
\okreview{\subsection{Proper names database}}
For our experimentation, we used a database of proper names\footnote{Company
address book}
which consisted of \okreview{$1,63,600$} entries, majority of which are Indian
names.
Majority of the names were made up of two parts - a first name and a
second name\footnote{We did not distinguish between a first name and a surname
for the set of experiments conducted. We believe that the
proposed algorithm shows a better performance if we create separate basis for
first names and surnames.}.
The first and surnames are considered as two different names and the duplicates
are removed.
So, to create a transcription dictionary one had to achieve transcription of
these unique names.
To test the performance of the proposed algorithm we further processed these
unique names
by removing names with two or less number of characters.
This resulted in a set of $|{\N}|$ = \okreview{25884} unique names.
\okreview{\subsection{Basis construction}}
The following results are obtained by using Algorithm 1 for the construction of
basis.
The names are first sorted in the descending order of their frequency of
occurrence in the entire database.
All the names whose frequency is
greater than or equal to \okreview{40\%} of the maximum frequency are taken as
${\B}_{init}$
which resulted in $|{\B}_{init}|$ = \okreview{225} (Step 1; Algorithm 1).
Using \texttt{isOrtho()} and \texttt{makeOrtho()}, ${\B}_{init}$ is checked for
its orthogonality
(see Appendix A)
i.e., the words in ${\B}_{init}$ that can be constructed from the other words
in 
${\B}_{init}$ are removed from it. The process of orthogonalization of basis is
described in Appendix C. This resulted in $|{\B}_{init}|$ = \okreview{224}.
This set is not strictly a basis because it doesn't satisfy the Spanning
Property (Property 2, Appendix A).
The unique names in the database are then constructed by joining the words in
${\B}_{init}$.
For a given name in the database ${\N}_{p}$, the set of all names in the basis
${\B}_{init}$,
which are the sub-words of ${\N}_{p}$, ${\B}_{p} = \{b_{p1}, b_{p2},
...,b_{pk}, ..., b_{p|{\B}_{p}|}\}$
is first collected (Step 7, Algorithm 1). Many of the names in ${\N}$ had their
substrings set, ${\B}_{p}$ empty.
This is because the initial basis contains only $224$ words out of which all
are `complete names'.
The probability of their occurrence as a part of other names is hence very low.
This resulted in many new words being appended to the initial basis
${\B}_{init}$ which resulted
in $|{\B}_{1}| = 25476$.
Using \texttt{isOrtho()} and \texttt{makeOrtho()} $|{\B}_{1}|$ is made
orthogonal,
and this reduced the size of  $|{\B}_{1}|$ to $10435$.
A significant growth and reduction in the size of basis is observed in this
iteration.
The above process of growing and pruning (orthogonalization) of the basis is
repeated
for a few iterations till no significant growth or reduction in the size of
basis is observed.
\\
For the set of experiments conducted, we chose the weight set as
\okreview{$\Lambda_{1} = \{0.4, 0.2, 0.1, 0.3\}$.
These values satisfy Equation (\okreview{17}). We chose this weight set as
follows. The basis is constructed
for 256 combinations of weights of the features of the cost function (16),
each weight taking a value from $[0, 0.1, 0.2, \cdots, 0.9, 1]$.
The chosen weight set is the one which resulted in the minimal overall cost
given by Equation (\okreview{(3)}.}

\begin{table}
\caption{$|{\B}|$ vs $|{\J}|$, variation of ${\C}$}
\centering
\begin{tabular}{c c c c c c}
\hline\hline
I. & $|{\B}_{m}|$ & $|{\B}|$ & $|{\J}|$ & $|{\B}_{m}||{\J}|$ & $|{\C}|$ \\
[0.4ex]
No &  &  &  & *$10^{-8}$ & \\ [0.4ex]
\hline
1 & 25476 & 10435 & 27614 & 7.03 &  23006\\
2 & 11131  &6168 &38570 & 4.29 &  16307.7\\
3 & 6549   &5985 &39629  &2.59 &  15348.1\\
4 & 6064   &5990 &39654  &2.40   &15326.1\\
5 & 6053   &5991 &39654  &2.40   &15326.1\\
\hline
\end{tabular}
\label{tab:basis_construction}
\end{table}

\begin{figure}
\centering
\includegraphics[width=0.45\textwidth]{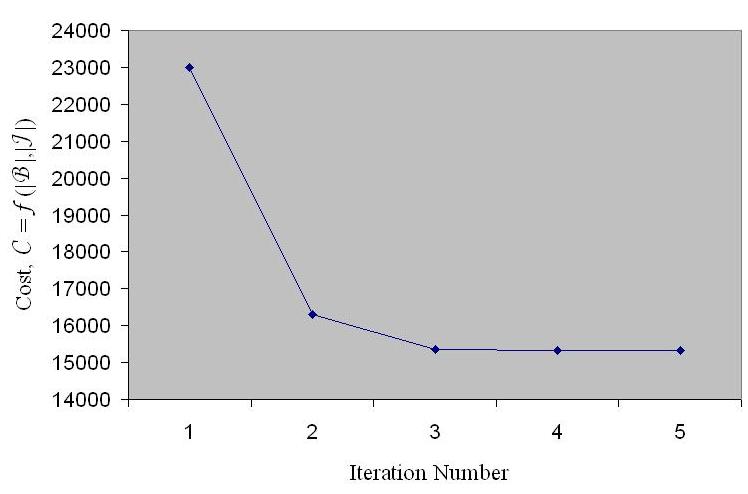}
\caption{Variation of $|{\C}|$ in Equation (3) over 6 iterations }
\label{fig:cost}
\end{figure}

\begin{figure}[t]
\centering
\includegraphics[width=0.45\textwidth]{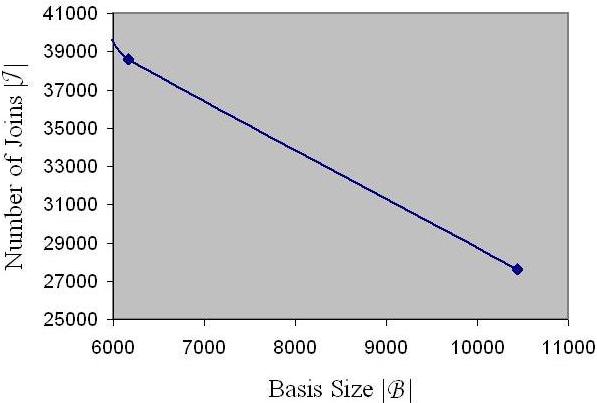}
\caption{Plot between basis size (${|\B|}$) and total number of joins
(${|\J|}$)}
\label{fig:BvsJ}
\end{figure}

\begin{figure}
\centering
\includegraphics[width=0.45\textwidth]{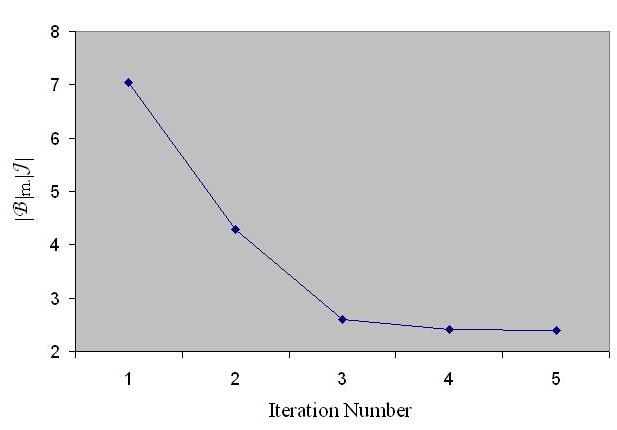}
\caption{Variation of $|{\B}_{m}||{\J}|$ over 5 iterations }
\label{fig:bm_j}
\end{figure}

Recall the cost function defined in Equation \ref{eq:cost} which has to be
minimized.
Table 1 shows the basis size before and after orthogonalization (columns 2 and
3 respectively),
the number of joins corresponding to the rank deficient basis (column 2) and
the corresponding cost ${\C}$.
Figure \ref{fig:cost} gives the actual variation of cost function $|{\C}|$ over
$5$
iterations of constructing the basis until convergence.
Figure \ref{fig:BvsJ} shows the plot between $|{\B}|$ and $|{\J}|$ for the
experimentation
performed which resembles Figure \ref{fig:BJ}.
We observe from Table \ref{tab:basis_construction} that the conditions
(\ref{cond1}) and (\ref{cond2})
derived for the convergence of the cost function
are met in the experimentation.
We see from Column 3 of Table \ref{tab:basis_construction} that the size of
basis is decreasing over iterations
which satisfies condition (\ref{cond1}).
We see from Column 5 of Table 1 that the product of basis size and number of
joins
($|{\B}_{m}||{\J}|$) is also reducing with iterations.
The variation of the product of $|{\B}_{m}|$ and $|{\J}|$ until convergence is
shown in Figure \ref{fig:bm_j}.

Compared to Algorithm 1, Algorithm 2 is simpler but computationally intensive.
The results obtained by using Algorithm 2 are as follows.
We chose $\Lambda_{2} = \{0.4, 0.3, 0.3, 0\}$ based on experimentation.
These values satisfy Equation (21). The obtained values for basis size and
number
of joins are $|{\B}|$ = {7174} and $|{\J}|$ = {38213}. The results presented
for both the algorithms
are obtained by not considering the feature $SA_{avg}$ in the cost function.

\okreview{
\subsection{Transcription of the optimal basis and the development of
Pronunciation Lexicon}

Now the obtained basis set has to be transcribed manually.
We followed the following process which is specific to Indian languages.
The basis words are written in Devanagari\footnote{a phonetic script}
(Hindi) script.
Using a lookup table, which maps Devanagari graphemes
to phonetic symbols, the Devanagari script of the basis words
is converted into
Festival TTS\cite{festival} accepted DARPA format and Microsoft
supported SAPI format.
For the sake of consistency we had three people transcribe
the basis separately and discuss among them to come up with a single set of
transcription for all the basis words.
One or more of these basis
transcriptions are concatenated to generate the transcriptions of the
proper names
in the names database. Table \ref{tab:basis} shows a few basis
words and their corresponding transcriptions in DARPA and SAPI formats while
Table \ref{tab:lexicon} shows some proper names from the name database,
their construction using the basis and the obtained phonetic
transcriptions using the proposed algorithms.
}

\begin{table}
\caption{A few Basis words and their transcriptions.}
\label{tab:basis}
\centering
\begin{tabular}{|c|l|l|l|l|}
\hline
No. & Basis & DARPA & SAPI\\ 
\hline\hline
1 & \texttt{kanth} & \texttt{k aa n th} & \texttt{k A n th}  \\
2 & \texttt{ma} & \texttt{m aa} & \texttt{m A}  \\
3 & \texttt{ra} & \texttt{r a} & \texttt{r a}  \\
4 & \texttt{je} & \texttt{jh ey} & \texttt{j E}  \\
5 & \texttt{shwar} & \texttt{s v ax r} & \texttt{S v a r}  \\
6 & \texttt{ram} & \texttt{r aa m} & \texttt{r A m}  \\
\hline
\end{tabular}
\end{table}

\begin{table*}
\caption{Some proper names, their construction using the basis words,
transcriptions in
DARPA and SAPI formats and Festival TTS G2P transcriptions of the names.}
\label{tab:lexicon}
\centering
\begin{tabular}{|l|l|l|l|l|}
\hline
Name & Word Seq& DARPA & SAPI & Festival (G2P rule base)  \\ 
\hline\hline
\texttt{ramakanth} & \texttt{ra ma kanth} & \texttt{r a m aa k aa n th} &
\texttt{r a m A k A n T h} & \texttt{r aa m ax k ae n th}\\
\texttt{rajeshwar} & \texttt{ra je shwar} & \texttt{r a jh ey s v ax r} &
\texttt{r a j E S v a r} & \texttt{r ax jh sh w ao r} \\
\texttt{narendra} & \texttt{na ren dra} & \texttt{n ax r ey n dh r ax} &
\texttt{n a r E n d r a} & \texttt{n r eh n d r ax}\\
\texttt{navyaram} & \texttt{navya ram} & \texttt{n ax v y aa r aa m} &
\texttt{n a w y a r A m} & \texttt{n ae v y aa r ae m}\\
\texttt{kamlesh} & \texttt{kam le sh} & \texttt{k ax m l ey sh} & \texttt{k
a m l E S} & \texttt{k ae m l ih sh} \\
\hline
\end{tabular}
\end{table*}

\okreview{\subsection{Accuracy of the system}
The next step is to ensure that the generated transcriptions obtained through
manual transcription of the basis words are correct.
The phonetic transcription of a proper name formed with two or more basis words
will vary when a basis word has multiple possible pronunciations.
This leads to a slight variation in the pronunciation of names.
Note that the higher the occurrence of a basis word
and the smaller its length, the greater is the possibility of
variation in its pronunciation.
The cost function formulated tries to maximize the length of the word
which goes into the basis,
thus minimizing the number of basis elements having multiple pronunciations.
Nevertheless, the optimization process results in
some basis words having multiple pronunciations which when used to
generate
transcriptions of names in the names database
result in a slight variation in the pronunciation of names.
They have to be taken care of manually when producing the final pronunciation
lexicon of the names in the database.
Our observation shows that the main source of multiple pronunciation
arise in the presence of a vowel.
While transcribing the basis words, sometimes,
one may not be aware whether the vowel in the basis word is long or short.
For example, in Table \ref{tab:lexicon},
the basis word \texttt{ra} in the first name \texttt{ramakanth} should
have a transcription \texttt{r aa} and in the second name \texttt{rajeshwar},
it should be transcribed as \texttt{r a}.
Using the same transcription of \texttt{ra} for both the names leads to a
slight variation in the pronunciation
of one of them. However, for longer basis words the probability of multiple
pronunciations is less.

After transcribing the basis, as mentioned above,
$200$ names from the proper names database are selected randomly
and their phonemic transcription is constructed by concatenating one or
more words using the transcribed optimal basis.
The obtained transcriptions are verified manually and the number of
correctly transcribed names are computed.
This shows that $85$\% of the names are correctly transcribed.
The inaccuracy in the transcriptions of the remaining $15$\% names is
due to the multiple pronunciations of
some basis words, majorly due to the following reasons:
\begin{enumerate}
\item [(a)] long vowels in the basis words transcribed as short vowels (and
vice versa),
\item [(b)] multiple pronunciations of phoneme \texttt{/s/} and \texttt{/t/}
which can have a phonetic form of [\texttt{/s/, /sh/}]
and [ \texttt{/t/,/T/} ] respectively.
\end{enumerate}
The above is the case when different names with different pronunciations are
spelled the same way.
Also note that different people spell the same name in different ways inspite
of having a unique
pronunciation for the name.
For example, an Indian name which has a phonetic spelling \texttt{ch au d a r
i} is
spelled in at least three different ways such as
\texttt{chaudary, chowdhari, chaudhari} depending on a person's choice.
Note that the proposed system generates slightly different transcriptions for
these three
instances of the same name.
}

\okreview{
\subsection{Comparison with Festival TTS G2P rule base}
Column $6$ in Table \ref{tab:lexicon} shows the phonetic transcriptions of the
names obtained using the
G2P rule base 
for the out of vocabulary (OOV) words used in the Festival TTS
engine\footnote{We used the Festival TTS G2P facility which was readily
accessible}.
The results show that the transcriptions generated using the
proposed method are found to be more accurate than the ones generated using the
Festival TTS rule base.
The proper names are also synthesized using these two kinds of transcriptions,
(a) one obtained using the process discussed in this paper and (b) the one
obtained using the G2P rule base used in Festival TTS,  with the Festival
TTS engine.
The names synthesized using the transcriptions obtained by the proposed method
are found to be
perceptually better. The perception test was carried out by asking two
persons who were not involved in the
basis transcription process to listen to the synthesized proper
names and rate the better of the two for each name (they had no
idea which transcription was used in the synthesizing process.}

{\em Note}: The algorithms proposed in this paper are
generic and are suitable for proper names of any language.
However, the system performs better if used for a database of
proper names of same origin or geographical area.
For example, the performance of the system is good when used to
transcribe a proper names database containing
only Indian names or only Chinese names, but degrades when used for a
database which has a mix of both Indian and Chinese proper names.
The results presented above are for a database containing a majority of
Indian names but not all.
Extending the same principle, if the system is used only for a database of
person names or place names and not a mixture of both,
the performance would be better.

\section{Conclusion}
Research on automatic G2P transcription has reported promising results for
phonetic transcription of regular text,
where G2P transcriptions follow certain rules. Generating phonetic
transcriptions of proper names,
where the general purpose G2P converter can not be applied directly, involves
human endeavor.
In this paper, an optimization approach for the automatic generation of
pronunciation
lexicon for proper names has been proposed.
We first construct a cost function and the
transcription problem reduces to one of
minimizing the constructed cost function.
Two algorithms for the identification of basis have been proposed and the
conditions for the convergence of the cost function have been derived.
Experimental results on real database of proper names validate the convergence
conditions derived and hence show that the developed optimization
framework helps in reducing the mundane task of transcribing proper names.
The formulated frame work is general and hence not restricted to
Indian proper names, though the experimentation has been carried out on an
Indian name database.
In fact, the framework is suitable for any database of proper names
irrespective of language.
Through experimental results we have demonstrated the working and the validity
of the proposed approach.

\appendices
\section{Basis}
In linear algebra, a basis ${\B}$ of a vector space ${\V}$, by definition,
is a set of linearly independent vectors that completely spans ${\V}$.
${\B} = \{b_{1}, ..., b_{m}\}$ is said to be a basis of vector space ${\V} =
\{v_{1}, ..., v_{n}\}$
if ${\B}$ has the following properties:
\begin{itemize}
\item
Linear independence property: If $a_{1}, ..., a_{m}$ are scalars and if
$a_{1}b_{1} + ... + a_{m}b_{m} = 0$, then necessarily $a_{1} = ... = a_{m} =
0$. 
This implies that $b_{1}, ..., b_{m}$ are orthogonal or $b_{1} \perp b_{2}
\perp ... \perp b_{m}$;
\item
Spanning property: For every $v_{k}$ in ${\V}$ it is possible to choose
scalars, 
$a_{1}, ..., a_{n}$ such that $v_{k} = a_{1}b_{1} + ... + a_{n}b_{n}$.
\end{itemize}

\section{Syntactic Rules}
It is advantageous to study/analyze the words syntactically before adding them
in to the basis for if the resultant basis element is not following any syntax,
its phonetic representation might not properly contribute to phonetically
represent a longer name which is a super set of it.
Syntactic knowledge is acquired by observing the sequences formed for a name.
Some rules are illustrated below.
V denotes a vowel and C denotes a consonant.
If the word is of a particular format, the following decisions would be taken
on its
candidature for the basis.
(Letters in bold represent the elements to be added to the basis).

\begin{itemize}
\item \textbf{CC  reject}\\
The pronunciation of a phone in a sequence of phones depends on the adjacent
phones.
Consonants depend on vowels for their pronunciation.
So, the basis element cannot be a pure consonant sequence.\\
Examples:
\begin{itemize}
\item shashank  sha+sha+\textbf{nk}
\item joseph - jose + \textbf{ph}
\item shantanu  sha + \textbf{nth} + anu
\item sunny - su+\textbf{nny}
\end{itemize}
All the words which are pure consonant strings are avoided. In other words, a
basis element must have at least one vowel.

\item \textbf{VC  OK}
\item \textbf{CV  avoid}
\item \textbf{VV  reject}

Introducing a split between two vowels is also not reasonable,
because most of the times, the combination of two vowel letters in English
forms a diphthongs.
They may be two characters but their combination is a single sound.
Example: shailendra - sh\textbf{a + i}lendra
\item Introducing a split between sh, th, dh also should be avoided  they are
two characters but their combination is a single phone/sound\\
Example: bharati  bhara\textbf{t + h}i
\end{itemize}

\section{Orthogonalization of Basis}
The following procedure is followed to make the basis orthogonal.
Names in the basis are sorted in descending order of their lengths.
For a word $b_{i}$ in the basis, a set of all words $b_{ik}$ which is a
substring of $b_{i}$ is collected,
${\B}_{i}$ = \{$b_{i1}, b_{i2}, ..., b_{ik}, ..., b_{|{\B}_{i}|}\}$.
If ${\B}_{i}$ is empty, $b_{i}$ is retained in the basis.
For the words whose ${\B}_{i}$ is not empty, elements of ${\B}_{i}$
are sorted in descending order of their lengths.
One name $b_{ik}$ from ${\B}_{i}$ is considered at a time and its position is
fixed in the word $b_{i}$.
The remainder of $b_{i}$ is filled with $b_{ik}$ in the order they appear in
${\B}_{i}$.
By the end of this process, $b_{i}$ must have been formed completely or
partially
with the available elements in ${\B}_{i}$. With one $b_{ik}$ at a time as the
first element to occupy its place in $b_{i}$, and filling the remainder of the
name with ${\B}_{i}$, we form $|{\B}_{i}|$ number of sequences for $b_{i}$.
If any one of the $|{\B}_{i}|$ sequences completely represents $b_{i}$,
then $b_{i}$ is deleted from ${\B}_{i}$; else it is retained.
The following example shows the sequences formed for the name, $b_{i}$ =
\texttt{krishna}

\noindent ${\B}_{i}$ = \{\texttt{krishn, krish, rish, kris, ris, ish, hna, na,
kr, hn, is, ri, sh}\}

\vspace{1ex}
\noindent\texttt{krishn \qquad partially constructed\\
krish na \qquad Fully constructed (1)\\
rish na \qquad Partially constructed\\
kris hna \qquad Fully constructed (2)\\
ris hna \qquad Partially constructed\\
kr ish na \qquad Fully constructed (3)\\
kris hna \qquad Fully constructed (4)\\
krish na \qquad Fully constructed (5)\\
kr ish na \qquad Fully constructed (6)\\
kris hn \qquad Partially constructed\\
kr is hna \qquad Fully constructed (7)\\
ri hna \qquad Partially constructed\\
kr sh na \qquad Partially constructed\\
}

In the above example, the word \texttt{krishna} in the existing basis can be
constructed in 7 different ways with the other existing basis elements.
So, it is not necessary to have it in the basis and hence removed from the
basis.

%
%
%
\section*{Acknowledgment}
%
%
The authors express their gratitude to Amol, Meghna
and Imran for their assistance in transcribing the
basis and evaluating the generated phonetic transcriptions.

\ifCLASSOPTIONcaptionsoff
  \newpage
\fi



\bibliographystyle{IEEEtran}
\bibliography{ref}
\end{document}